\documentclass[10pt,twocolumn,letterpaper]{article}

\usepackage{mmstyle}
\usepackage{cvpr}
\usepackage{times}
\usepackage{epsfig}
\usepackage{graphicx}
\usepackage{amsmath}
\usepackage{amssymb}
\usepackage{algorithm}
\usepackage{algorithmicx}
\usepackage{algpseudocode}
\usepackage{caption}
\usepackage{booktabs}  
\usepackage{threeparttable}  
\usepackage{multicol}  
\usepackage{multirow}  
\usepackage{tabu}  
\usepackage{float}

\usepackage[pagebackref=true,breaklinks=true,letterpaper=true,colorlinks,bookmarks=false]{hyperref}

\cvprfinalcopy 

\ifcvprfinal\fi
\begin{document}

\title{Density-Aware Graph for Deep Semi-Supervised Visual Recognition}

\author{Suichan Li$^{1,2}$\\
\and 
Bin Liu$^{1,2}$\\
\and 
Dongdong Chen$^{3}$\\
\and
Qi Chu$^{1,2}$\thanks{corresponding author.}\\
\and
Lu Yuan$^{3}$\\
\and
Nenghai Yu$^{1,2}$\\
\and
$^1$School of Information Science and Technology, University of Science and Technology of China\\
$^2$Key Laboratory of Electromagnetic Space Information, the Chinese Academy of Sciences\\
$^3$Microsoft Research
}

\maketitle
\thispagestyle{empty}

\begin{abstract}
Semi-supervised learning (SSL) has been extensively studied to improve the generalization ability of deep neural networks for visual recognition. To involve the unlabelled data, most existing SSL methods are based on common density-based cluster assumption: samples lying in the same high-density region are likely to belong to the same class, including the methods performing consistency regularization or generating pseudo-labels for the unlabelled images. Despite their impressive performance, we argue three limitations exist: 1) Though the density information is demonstrated to be an important clue, they all use it in an implicit way and have not exploited it in depth. 2) For feature learning, they often learn the feature embedding based on the single data sample and ignore the neighborhood information. 3) For label-propagation based pseudo-label generation, it is often done offline and difficult to be end-to-end trained with feature learning. Motivated by these limitations, this paper proposes to solve the SSL problem by building a novel density-aware graph, based on which the neighborhood information can be easily leveraged and the feature learning and label propagation can also be trained in an end-to-end way. Specifically, we first propose a new Density-aware Neighborhood Aggregation(DNA) module to learn more discriminative features by incorporating the neighborhood information in a density-aware manner. Then a novel Density-ascending Path based Label Propagation(DPLP) module is proposed to generate the pseudo-labels for unlabeled samples more efficiently according to the feature distribution characterized by density. Finally, the DNA module and DPLP module evolve and improve each other end-to-end. Extensive experiments demonstrate the effectiveness of the newly proposed density-aware graph based SSL framework and our approach can outperform current state-of-the-art methods by a large margin.
\end{abstract}
\section{Introduction}
\begin{figure}[t]
	\includegraphics[width=1.0\linewidth]{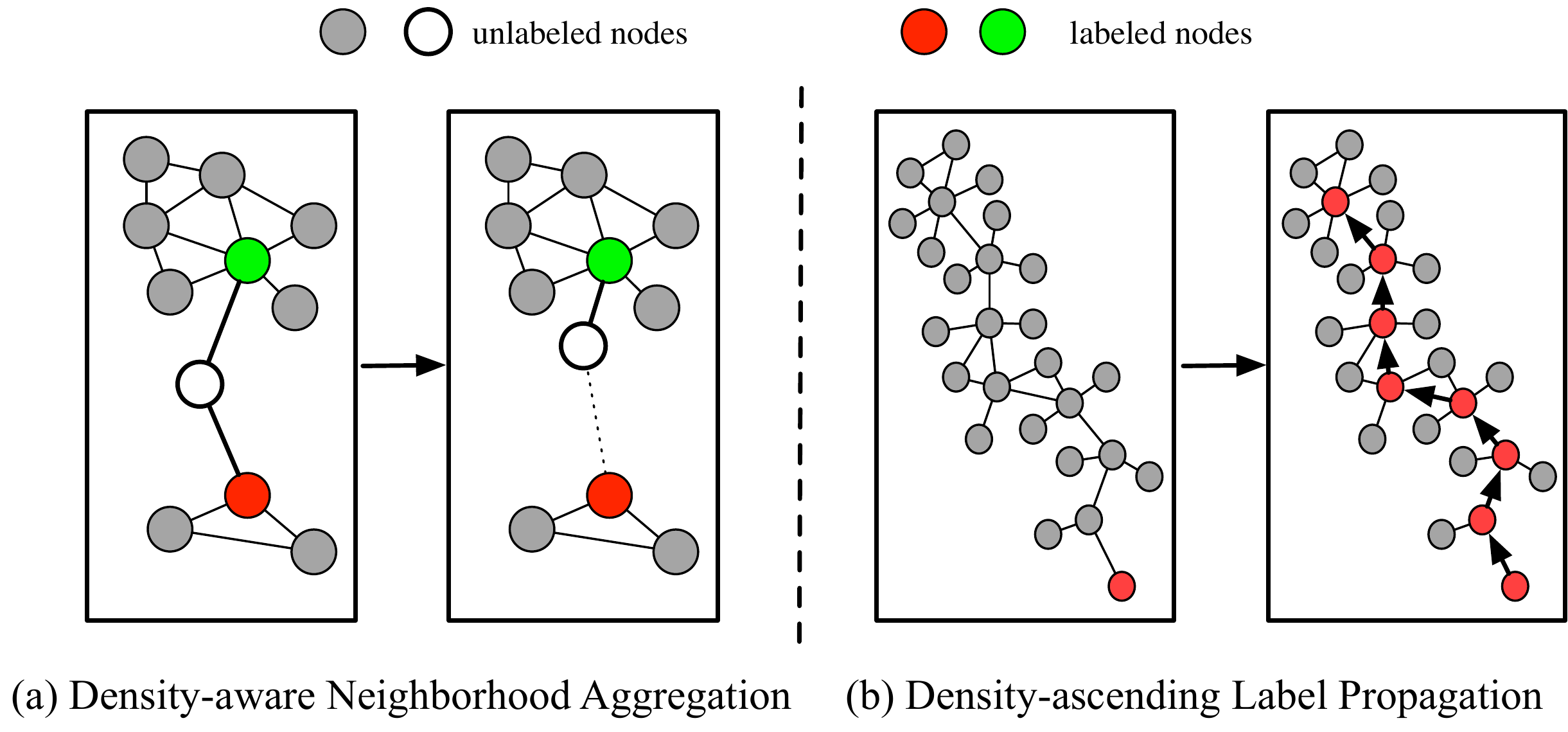}
	\caption{\small (a) Density-aware Neighborhood Aggregation(DNA) scheme prefer neighbors with higher density when target sample has equal similarity with the neighbors which belong to different clusters. (b) Density-ascending Path-based Label Propagation(DPLP) constructs a density-ascending path and propagates the labels from labelled samples to unlabelled samples within the path.} \label{fig:case}
 \vspace{-1em}
\end{figure}
Recently, semi-supervised learning (SSL) has been extensively studied to improve the generalization ability of deep neural networks for visual recognition by utilizing limited labelled images and massive unlabelled images. For leveraging the unlabelled data, current SSL methods  are usually based on a common density-based \textit{cluster assumption}, $\ie$ samples lying in the same high-density region are likely to belong to the same class\cite{chapelle2005semi,ICT2019}. Its equivalent one is \textit{low-density separation assumption}, which states that the decision boundary should not cross high density regions, but instead lies in low density regions. Current methods mainly focus on enforcing the low-density separation assumption by encouraging invariant prediction for perturbations around each unlabelled data point or same predictions for nearby samples, including consistency-regularization based methods~\cite{laine2016temporal,tarvainen2017mean,ICT2019,berthelot2019mixmatch} and pseudo-label based methods~\cite{lee2013pseudo,shi2018transductive,iscen2019label}. 

While \textit{density-peak assumption}~\cite{Rodriguez1492} states that high-density samples are more likely to be the center of a cluster, thus samples with high density can encode more representative information of the cluster, which are valuable clues for semi-supervised learning. However, current methods have not consider such density information explicitly or exploited it in depth. Moreover, as we know, the performance of current SSL frameworks is mainly determined by two aspects: feature learning and pseudo label generation for unlabelled samples. Nevertheless, when learning the feature embedding, current methods mainly leverage the single data sample and ignore the abundant neighborhood information which is helpful to learn more discriminative features. And for pseudo label generation, existing methods often directly choose current model predictions\cite{lee2013pseudo} or perform label propagation\cite{iscen2019label} to generate labels for unlabeled samples, which induce either inaccurate pseudo labels, or high matrix computation cost which is difficult to be end-to-end trained with the feature learning part.

Motivated by above observations, this paper proposes a novel density-aware graph based SSL framework. By building a density-aware graph, the neighborhood information can be easily utilized for each data sample. More importantly, the feature learning part and the label propagation part can also be end-to-end trained in the newly proposed framework. Further more, to better leverage the density information, we explicitly incorporate it into these two parts respectively. 

Specifically, given the labelled and unlabelled data samples, a density-aware graph will be first built and we define the density for each node in the graph. Then for feature embedding learning, rather than only based on each single data sample, we propose to aggregate the neighborhood features to enhance the target features instead, which is demonstrated to be very useful in modern Graph Neural Networks(GNNs)~\cite{scarselli2008graph,kipf2016semi,mne_iccv19}. However for the current aggregation schemes, the aggregation weights are often only characterized by the feature similarity between target node and its neighbors.
In such case, when the target node has equal similarity with two neighbors which belong to two different clusters, it will give the same weight to these two neighbors. Motivated by the aforementioned density-peak assumption, we propose a novel Density aware Neighborhood Aggregation(DNA) scheme. Concretely, besides considering the feature similarity, we take the neighbor density information into account as well when calculating the aggregating weights. Intuitively, we want higher density neighbors to have higher importance. One simple explanation of this strategy is illustrated in Fig.~\ref{fig:case}(a).

To generate pseudo-labels for unlabelled samples more efficiently, we follow the basic label-propagation scheme which propagates the labels from labelled samples to unlabelled samples. However, we are not going to perform label propagation through a linear system solver used in~\cite{iscen2019label,zhu2002learning}, which induces high matrix computational cost and works offline while training. Inspired by the aforementioned density assumption again, given one unlabelled sample, we argue the pseudo-label generated from neighboring samples with higher density is more possibly precise than that from neighbors with lower density. Based on this insight, we further propose a novel Density-ascending Path-based Label Propagation (DPLP) module. Specifically, for each sample, we will construct a density-ascending path where densities of samples are characterized by ascending-order, and perform label propagation within this path, which is efficient and can be online trained with feature learning in an end-to-end fashion. A graphical illustration can be found in Fig.~\ref{fig:case}(b). 

In summary, the main contributions of this work include: (1) We propose a novel density-aware graph based SSL framework. To the best of our knowledge, this is the first work which exploits the density information explicitly for deep semi-supervised visual recognition; (2) A new Density-aware Neighborhood Aggregation module and Density-ascending Path-based Label Propagation module are designed for better feature learning and pseudo label generation respectively. The two modules are integrated into a unified framework and can be end-to-end trained; (3) Extensive experiments demonstrate the effectiveness of the proposed framework, which significantly outperforms the current state-of-the-art methods.
\section{Related Works}
\noindent \textbf{Consistency-regularization for SSL.}
These methods usually apply a consistency loss on the unlabeled data, which enforce invariant predictions for perturbations of unlabelled data. For example, $\Pi$-model~\cite{laine2016temporal} proposes to use a consistency loss between the outputs of a network on random perturbations of the same image, while Laine $\etal$~\cite{laine2016temporal} apply consistency constraints between the output of the current network and the temporal average of outputs during training. The mean teacher (MT) method ~\cite{tarvainen2017mean} replaces output averaging by averaging of network parameters. To utilize the structural information among unlabeled data points, \cite{shi2018transductive} applies a Min-Max Feature regularization loss to encourage networks to learn features with better between-class separability and within-class compactness. Similarly, Luo $\etal$~\cite{luo2018smooth} utilize the contrastive loss to enforce neighboring points to have consistent predictions while the non-neighbors are pushed apart from each other. Although these methods have exploited neighborhood and density information, they are in the form of regularization terms or loss functions. By contrast, our method proposes to aggregate neighborhood features to enhance the target feature in a more explicit density-aware manner.

\noindent \textbf{Pseudo-labeling for SSL.} To leverage unlabelled data, pseudo-label based methods try to assign \textit{pseudo labels} to the unlabeled samples based on labelled samples, then train the network in a fully supervised way. To generate precise pseudo labels, Lee $\etal$~\cite{lee2013pseudo} use the current network predictions with high confidence as pseudo-labels for unlabeled examples. Shi $\etal$~\cite{shi2018transductive} use the network class prediction as hard labels for the unlabeled samples and introduce an uncertainty weight. Recently, Iscen $\etal$~\cite{iscen2019label} employ graph-based label propagation to infer labels for unlabeled samples. However, they perform label propagation through a linear system solver on the training set offline with high computational cost, thus cannot be trained in an end-to-end way. In this work, we propose to construct a density-ascending path and perform label propagation within this path, which is much more efficient and can be end-to-end trained.

\noindent \textbf{Neighborhood Aggregation in GNNs.}
Modern GNNs broadly follow a neighborhood aggregation scheme, where each node aggregates feature vectors of its neighbors to get more representative feature vector~\cite{xu2018how}. Different GNNs can vary in how they perform neighborhood aggregation. For example, Kipf $\etal$~\cite{kipf2016semi} use mean-pooling based neighborhood aggregation and Hamilton $\etal$~\cite{hamilton2017inductive} propose three aggregator functions: Mean aggregator, Max-Pooling aggregator and LSTM aggregator. Recently, inspired by self-attention mechanism, Petar $\etal$~\cite{velivckovic2017graph} propose an attention-based aggregation architecture by learning adaptive aggregation weights. Li $\etal$~\cite{mne_iccv19} extend the attention-based aggregation by supervising the attention weights with node-wise class relationship. After careful study, we find most of these neighborhood aggregation methods only consider the feature similarity between target sample and its neighbors when defining the aggregating weights. However, density information is shown to be a very important clue for SSL. Therefore, besides feature similarity, this paper also takes the neighborhood density into consideration and proposes a novel density-aware neighborhood aggregation scheme.
\section{Preliminary}
In semi-supervised learning, a small amount of labelled training samples $\mathcal{D}_{l}=\{(x_i,y_i)\}_{i=1}^l$ and a large set of unlabelled training samples $\mathcal{D}_u = \{x_j\}_{j=1}^u$ are often given, where $l$ and $u$ are number of labelled and unlabelled samples respectively and usually $l \ll u$. Then the goal of SSL is to leverage both $\mathcal{D}_{l}$ and $\mathcal{D}_u$ to train a better and generalized recognition model. Formally, let $m = l + u $ be the total number of training samples, $f_{\theta}$ be the feature extractor and $h_{\phi}$ be the classifier, current deep SSL methods adopt a similar optimization formulation:
\begin{equation}
\underset{\theta,\phi}{\text{min}}  \sum_{i=1}^{m} \ell \left( h_{\phi}(f_{\theta}(x_i)), \hat{y}_i\right) + \lambda R(\theta,\phi,\mathcal{D}_l,\mathcal{D}_u)
\label{eq:optimizer}
\end{equation}
where $\ell$ is the loss function like cross-entropy loss. For labelled data, $\hat{y}$ is the ground-truth label, while for unlabelled data, $\hat{y} $ can be pseudo-label. $R$ is the regularization term, which encourages the model to generalize better to unseen data. Inspired by~\cite{grandvalet2005semi, NIPS1991_440}, we add regularization term as follow:
\begin{equation}
 \small 
\begin{aligned}
    R & =-\frac{1}{m}\sum_{i=1}^{m}\sum_{j=1}^{n_c}h_{j}(x_i;\theta,\phi)logh_{j}(x_i;\theta,\phi) \\
    & -  \frac{1}{n_c} \sum_{j=1}^{n_c}log\bar{h}_{j}(X;\theta,\phi) 
\end{aligned}
\end{equation}
where $n_c$ is the number of classes, $\bar{h}_{j}(X;\theta,\phi)$ represents the mean softmax predictions of the model for category $j$ across current training batch. The first term is the entropy minimization objective defined in~\cite{grandvalet2005semi}, which simply encourages the model output to have low entropy; while the second term encourages the model to predict each class with equal frequency on average~\cite{NIPS1991_440}.

\section{Method}

 \begin{figure*}[t]
	\centering
	\includegraphics[width=0.9\linewidth]{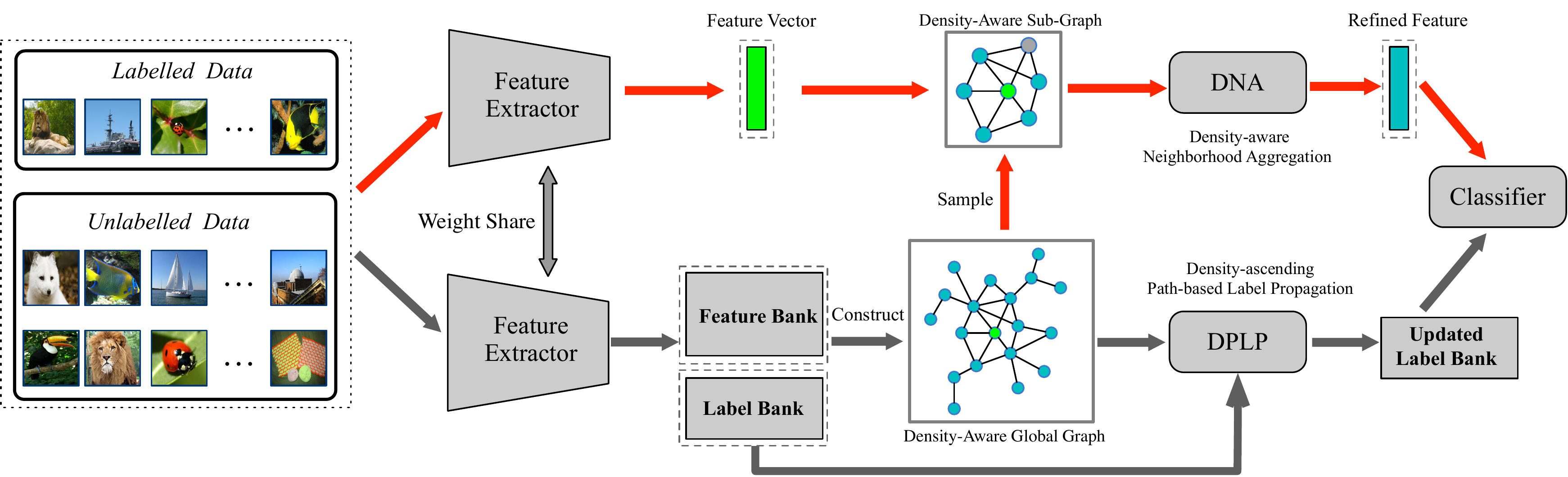}
	\caption{\small The overview of the proposed density-aware graph based SSL framework. The red line demonstrates the data flow in training, which includes feature extraction, sub-graph sampling and density-ware neighbourhood aggregation. The grey line shows the flow for density-ascending path-based label propagation. }
	\label{fig:pipeline}
	\vspace{-1em}
\end{figure*}

\noindent\textbf{Overview.} In this work, we introduce a unified framework for joint feature learning and label propagation on a density-aware graph for semi-supervised visual recognition, which can be trained in an end-to-end fashion. A graphical overview of the proposed framework is depicted in Fig.\ref{fig:pipeline}. First, we construct a $k$-nearest neighbor graph and define the \textit{density} for each node in the graph. Then based on the density-aware graph, we propose to learn the feature embedding and pseudo label generation simultaneously for each node in the graph. Specifically, for each target node, we will sample its neighborhood sub-graph and learn feature embedding on this sub-graph by incorporating the neighborhood information with Density-aware Neighborhood Aggregation(DNA). For pseudo label generation, we propose Density-ascending Path-based Label Propagation (DPLP), $\ie$, build a density-ascending path for each node in the graph and propagate the labels from labelled nodes to unlabelled nodes within this path.

\subsection{Density-Aware Graph}

Given a pre-trained feature extractor and classifier, we first extract the feature vectors and label predictions for all training samples, and organize them as Feature Bank and Label Bank respectively, which can be accessed through index later. Based on the features in the feature bank, we construct the global $k$-nearest neighbor affinity graph ($k$ = 64 in this work). We then define the \textit{density} $\rho$ for each node $u$ in the graph as:
\begin{equation}
\rho_{u}=\frac{1}{\left|\mathcal{N}_{k}(u)\right|} \sum_{v \in \mathcal{N}_{k}(u)} \mathbf{\tilde{f}}_{u}^{T} \mathbf{\tilde{f}}_{v}  \label{Eq:density}
\end{equation}
where $\mathcal{N}_{k}(u)$ is the $k$-nearest neighbors of node $u$, and $\mathbf{\tilde{f}}_u$, $\mathbf{\tilde{f}}_v$ are the L2-normalized feature embedding of node $u$ and $v$. Intuitively, this formula expresses the density of each node $u$ as the average of the similarities between $u$  and its neighbors. Need to note that other definitions of the density can also be considered, such as the number of neighbors whose similarity with target node is greater than a predefined threshold~\cite{Rodriguez1492}. But it is not the focus of this work. We refer to the graph equipped with the density as \textit{Density-Aware Graph}; 

\subsection{Density-Aware Neighborhood Aggregation}
Current methods mainly focus on regularizing the output to be smooth near a data sample locally but learn the feature embedding only based on each single data sample. That is to say, they have not fully explored the important neighborhood information for feature learning, which is demonstrated to be very useful in other tasks \cite{Zhuang_2019_ICCV, mne_iccv19, Sabokrou_2019_ICCV, han2019once,zhou2019dup}. Motivated by this observation, we propose to enhance the feature embedding of each target sample by aggregating the neighborhood features. Specifically, for each sample in the current training batch, we first pass it through the backbone to get the feature vector, and get corresponding node in the global density-aware graph (referred as target node), then we sample the neighborhood nodes of the target node from the global graph to obtain a sub-graph.  

\noindent \textbf{Sub-Graph Construction}
For construction of the sub-graph, we follow~\cite{mne_iccv19} and organize the neighbor nodes as a Tree-Graph. In particular, we take the target node as the root node, then build the tree in an iterative fashion.  Each time, we extend all the leaf nodes by adding their $k$ nearest neighbors from global-graph as the new leaf nodes. The tree graph grows until it reaches a predefined depth $h$. Based on Tree-Graph, we then iteratively perform feature aggregation among connected nodes and gradually propagates information within the sub-graph from leaf nodes to the target node. 
In the experimental part, we will study the effect of the number of sampling neighbors $k$ and graph depth $h$. 

\noindent \textbf{Density-aware Neighborhood Aggregation(DNA)}
\label{sec:dna}
After sampling of sub-graph, we propose to improve target feature embedding by aggregating its neighbors embedding in the sub-graph. General aggregation strategies like mean-pooling and max-pooling cannot determine which neighbors are more important. Recently, to adaptively aggregate the features, \cite{velivckovic2017graph, mne_iccv19} proposed an attention-based architecture to perform neighboring aggregation, whose aggregation weights are characterized by the feature similarity between target node and its neighbors. Formally, the adaptive aggregation can be denoted as:
 \begin{equation}
\mathbf{f}_{u}^{\prime}=\mathbf{W}_{A}\left(\mathbf{f}_{u}+\sum_{v \in \mathcal{N}(u)} a_{u, v} \mathbf{f}_{v}\right)+\mathbf{b}_{A}
\end{equation}
where $\mathbf{W}_{A}, \mathbf{b}_{A}$ are extra parameters for feature transformation. 
And $a_{u,v}$ is the aggregation weight denoting how much neighbor node $v$ contributes to the target node $u$:
\begin{equation}
a_{u,v} =  \frac{e^{p_{u,v}} } {\sum_{k \in \mathcal{N}(u)} e^{p_{u,k}}},   \label{Eq:aggregation_weight}
\end{equation}
\begin{equation}
p_{u,v} =\mathbf{\tilde{f}}^T_{u} \mathbf{\tilde{f}}_{v}, \quad
 \mathbf{\tilde{f}}_{u} =  \frac{\mathbf{f}_{u}}{\|\mathbf{f}_{u}\|}, \quad  
 \mathbf{\tilde{f}}_{v} =  \frac{\mathbf{f}_{v}}{\|\mathbf{f}_{v}\|}
\end{equation}

In details, we first perform feature L2-normalization, then define the similarity $p_{u,v}$ as the inner product of L2-normalized feature, and get the final aggregation weights by normalizing the similarity with the softmax operation.

However in SSL, we find only considering the aggregation weight with feature similarity is sub-optimal. In this way, if the target node has equal similarity with two neighbors that belong to two different clusters, the same aggregation weights will be assigned to these two neighbors. In fact, based on the \textit{density-peak assumption}, the nodes with higher density are more closer to the cluster center and more discriminative. Therefore, besides the feature similarity information, we propose to incorporate the density information of each neighbor as well when calculating the aggregation weights. By default, we simply combine feature similarity $\mathbf{p}$ and density $\mathbf{\rho}$ with element-wise summation and rewrite Eq.~\ref{Eq:aggregation_weight} as follows:
\begin{equation}
   a_{u,v} = \frac{ e^{(p_{u,v} + \rho_{v})}} { \sum_{k \in \mathcal{N}(u)} e^{(p_{u,k} + \rho_{k})}}.  \label{Eq:attention_weight_new}
\end{equation}

\subsection{Density-ascending Path for Label Propagation}
\label{sec:dlp}
\noindent \textbf{Density-Ascending Path Construction.} We construct the density-ascending path for each node in the global density-aware graph. More specifically, for node $u$, we initialize the density-path as one-element set $\{u\}$. Then we add one new nearest neighbor node $v$, whose density is greater than the previous added node. We iteratively perform this process until the distance between the candidate node and the last added node is greater than a predefined threshold.

For notation clarity, we define the Density-Ascending Path as $\mathcal{P}(u)=\{v_1, v_2, ..., v_k,...\}$, where $v_k$ is the node added to the path at $k$-$th$ step. Supposing the added node at $k$-$th$ step is $v_k$, then node to be added is the neighbor node $v_{k+1}$ with higher density:
\begin{equation}
v_{k+1} = \text{arg} ~ \underset{v}{\text{min}} ~  \Psi ( \mathbf{f}_{v_{k}}, \mathbf{f}_{v}  ), \\
v \in \{ w| \rho_{w} > \rho_{v_{k}} \}   \label{Eq:density_path}
\end{equation}
where $\Psi(\cdot)$ is a distance metric function, and we choose the L2-Euclidean distance metric in this work by default, i.e., 
$\Psi ( \mathbf{f}_{v_{k}}, \mathbf{f}_{v}  ) = \left \| \mathbf{f}_{v} - \mathbf{f}_{u} \right \|_2$
To alleviate the influence of irrelevant neighbors, we define a threshold $\sigma$, to terminate the growth of the density-ascending path, i.e., for each node pair ($v_{k}$,$v_{k+1}$) in $\mathcal{P}(u)$, it satisfies: $\left \| \mathbf{f}_{v_{k}} - \mathbf{f}_{u_{k+1}} \right \|_2 \leq \sigma$. 
\begin{algorithm}[t]
\small
 \caption{\small{Density-Ascending Path-based Label Propagation}}
 \renewcommand{\algorithmicrequire}{\textbf{Input:}}
 \renewcommand{\algorithmicensure}{\textbf{Output:}}
 \begin{algorithmic}[1]
 \Require Density-Aware Graph $\cG$, Label Bank $\cB$, labeled node indices $\cI_{l}$,  unlabeled node indices $\cI_{u}$.
 \Ensure Updated label Bank $\cB^{'}$
 \State $\cI^{'}_{u} = \cI_{u}$,  $\cB^{'} = \cB$
  \vspace{0.3em}
  \State $\cI^{'}_{l} = \Call{SortInDensityDecending}{\cI_{l},  \cG}$
  \For{$i \in \cI^{'}_{l}$}
  \State $\cP = \Call{ConstructDensityAscendingPath}{i,  \cG}$
  \For{$ j \in \cP$}
  \If{$ j  \in \cI^{'}_{u}$}
   \State$ \cB^{'} = \Call{UpdateLabelBank}{\cB^{'}, i, j }$
   \State $\cI^{'}_{u} = \cI^{'}_{u} \setminus  \{j\}$
  \EndIf
  \EndFor
  \EndFor
  \For{$i \in \cI^{'}_{u}$}
  \State $\cP = \Call{ConstructDensityAscendingPath}{i,  \cG}$
  \For{$ j \in \cP$}
  \If{$ j  \in \cI^{'}_{l}$}
   \State $\cB^{'} =\Call{Updatelabelbank}{\cB^{'}, j, i}$
  \EndIf
  \EndFor
   \EndFor
  \State \Return $\cB^{'} $
 \end{algorithmic}
 \label{alg:density-path}
\end{algorithm}

\begin{algorithm}[t]
\small
 \caption{\small{Density-Aware Graph for Deep SSL}}
 \begin{algorithmic}[1]
 \Require \small{Training set $\{ \mathcal{X}, \mathcal{Y}_l\}$}, training epochs $T$, training iterations $J$, initial feature bank $\mathcal{F}^{0}$, initial label bank $\mathcal{L}^{0}$, labeled indices $\cI_{l}$, unlabeled indices $\cI_{u}$ 
 \Require \small{Feature Extractor $f_\theta$,  Classifier $g_\phi$}
 \Require \small{DensityAwareNeighborhoodAggregation $\Call{Dna}{\cdot|\omega}$ }
 \Require \small{DensityAscendingPathLabelPropagation $\Call{Dplp}{\cdot}$ }
  \For{$t \in \{1,\dots,T\}$}
  \State $\mathcal{F}^{t}, \mathcal{L}^{t} = \Call{Initalize}{f_\theta,g_\phi, \mathcal{X}, \mathcal{Y}_l, \mathcal{F}^{t-1}, \mathcal{L}^{t-1}}$
  \State $\mathcal{G}^t = \Call{ConstructDensityAwareGraph}{\mathcal{F}^{t}, \mathcal{L}^{t}}$
  \State$\cY = \Call{Dplp}{\cG^t,\cL^t,\cI_{l}, \cI_{u}}$  \Comment{ label propagation}
 \For{$ i \in \{ 1,\dots, J \}$}
 \State$ X^i, Y^i =  \Call{GetTrainingBatch}{\cX, \cY, i}$
 \State$F^i = f(X^i;\theta)$
  \State$\cG_{s} = \Call{SampleSubGraph}{F^i, \cG^t}$
 \State$F^i = \Call{Dna}{F^i, \mathcal{G}_{s};\omega}$ \Comment{feature aggregation} 
 \State$P^i = g(F^i;\phi)$     \Comment{ category predictions} 
 \State$\ell = \Call{CaculateLoss}{P^i,Y^i}$
 \State$\Call{UpdateParameter}{\ell, \theta, \phi, \omega}$ \Comment{back-propagation}
 \EndFor
 \EndFor
 \end{algorithmic}
 \label{alg:Density_Arare_Graph}
\end{algorithm}
Before entering the Density-ascending Path-based Label Propagation, we first introduce the following assumption.

\vspace{0.5em}
\noindent \textbf{Assumption:}  \textit{The labelled nodes with higher density in the density-ascending path are more possible to provide correct pseudo labels than the ones with lower density. }

\noindent \textbf{Explanation:} As stated in the \textit{cluster assumption}, samples with the same label are more likely to lie in the high-density region. Meanwhile, the \textit{density-peak assumption}~\cite{Rodriguez1492} shows that high-density nodes are more likely to be the center of a cluster. Thus for one unlabelled node, the labelled nodes with higher density are more representative and more likely to provide correct pseudo labels than the ones having lower density in the same density-ascending path. 

Based on \textit{density-ascending path}, now we introduce our Density-ascending Path-based Label Propagation(DPLP) algorithm. Specifically, we first sort all the labelled nodes based on the density in descending order, then for each labelled node, we construct a density-ascending path and use the label of the max-density labelled node to update the entries of Label Bank corresponding to all the unlabelled nodes in this path. For remaining unlabelled nodes,  we also construct a density-ascending path for each of them and update the corresponding entry of Label Bank using the label of labeled node with highest density in the path. The detailed procedures are summarized in Alg.~\ref{alg:density-path}.

\subsection{Density Aware Graph-based SSL Framework}
The above Density-aware Neighborhood Aggregation and Density-ascending Path-based Label Propagation are integrated into a unified Density Aware Graph-based SSL framework (dubbed as ``DAG") and can be trained in an end-to-end fashion. We summarize the whole training process in Alg.~\ref{alg:Density_Arare_Graph}. Specifically, at 
 the beginning of each epoch, we update the Feature Bank and Label Bank with the latest feature extractor and classifier. Then construct a new global density-aware graph based on current Feature Bank and perform density-ascending path label propagation based on DAG and current Label Bank. Need to note that we always use ground-truth labels for the Label Bank entries of labeled samples. 
After the above training preparation, we then start to train the framework by sampling batch images and labels. In details, for each batch of images, we feed them into the feature extractor and enhance their output features by density-aware neighborhood aggregation on sub-graphs before feeding them into the classifier.
\section{Experiments}
\begin{table}[t]
  \begin{center}
  \setlength{\tabcolsep}{3.5mm}{
    \begin{tabular}{lcccc}  
    \toprule  
   {\textbf{Method}}&  
    \multicolumn{1}{c}{CIFAR10}&\multicolumn{1}{c}{CIFAR100}\cr  
    No. of labelled images &1000&4000\cr  
    \hline \hline
    Baseline & 10.96  & 43.34      \cr
    NA w/o density& 9.46 &  40.76 \cr  
   NA with density &\textbf{9.18}  &  \textbf{40.33} \cr  
    \bottomrule  
    \end{tabular} }
    \captionof{table}{\small Effectiveness of Neighborhood Aggregation(NA) on CIFAR10 and CIFAR100. } \label{tab:Neighborhood_Aggregation}
     \end{center}
    \vspace{-2em}
\end{table}
\begin{figure}[t]
	\centering
	\includegraphics[width=1.0\linewidth]{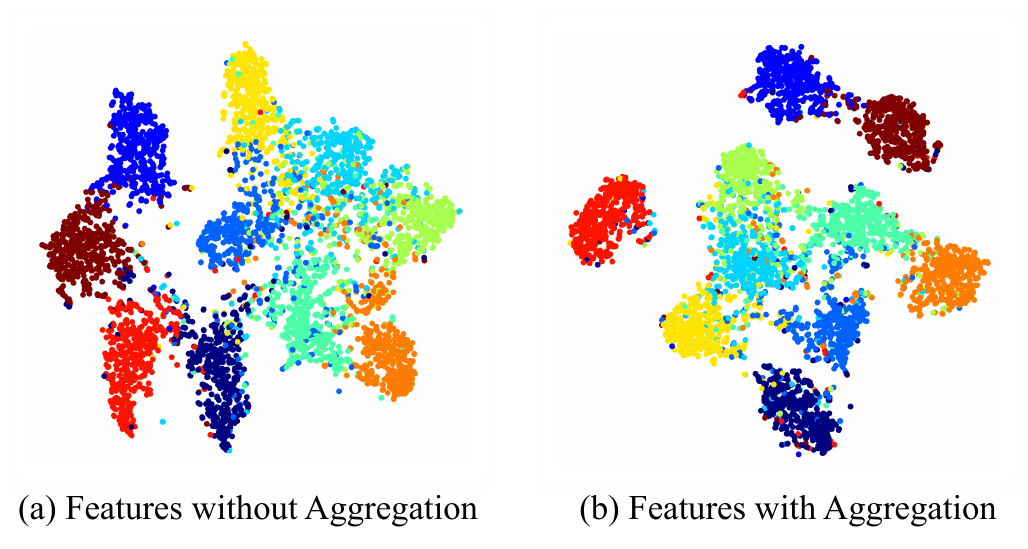}
	\caption{\small Visualization of feature embeddings on CIFAR10. Each dot in the figure corresponds to one image, and different colors represent different classes. (Best viewed in color.)} \label{fig:feat_visualize}
	 \vspace{-1em}
\end{figure}
\subsection{Experimental Setup}
\noindent \textbf{Dataset Setup.} To verify the effectiveness, we conduct experiments on three popular datasets, namely CIFAR10~\cite{krizhevsky2009learning}, CIFAR100~\cite{krizhevsky2009learning} and Mini-ImageNet~\cite{vinyals2016matching}.
In details, CIFAR10 and CIFAR100 both contain 50k images for training and 10k images for testing with resolution $32\times32$, but coming from 10 and 100 classes respectively. Following the standard SSL setting, for CIFAR10, we perform experiments with 50, 100, 200, and 400 labelled images per classes.
And for CIFAR100, we experiment with 40 and 100 labelled images per class. 
For Mini-ImageNet, it is a subset of ImageNet\cite{deng2009imagenet} and consists of 100 classes with 600 images per class of resolution $84\times84$. With the same setting as~\cite{iscen2019label}, we randomly assign 500 images from each class to the training set, and 100 images to the test set. The train and test sets therefore contain 50k and 10k images. We then experiment with 40 and 100 labelled images per class for SSL. 
We perform ablation study on CIFAR10 and CIFAR100 with an independent validation set of 5K instances sampled from the training set as~\cite{athiwaratkun2018there,oliver2018realistic} and compare with state-of-the-art methods on the standard test set.

\noindent \textbf{Implementations.} 
For model architecture, we use the same 13-layer CNN network as in ~\cite{laine2016temporal,tarvainen2017mean,iscen2019label} for CIFAR10/100 and ResNet-18~\cite{he2016deep} for Mini-ImageNet as~\cite{iscen2019label}. The SGD optimizer is used with the initial learning rate  0.1 and 0.2 for CIFAR10/100 and Mini-ImageNet respectively. We decay the learning rates by 0.1 at 250 and 350 epochs and obtain the final model after 400 epochs. We augment training images with random cropping and horizontal flipping as \cite{laine2016temporal,tarvainen2017mean}. Inspired by~\cite{zhang2018mixup,ICT2019,berthelot2019mixmatch}, we also employ Mixup strategy\cite{zhang2018mixup} to augment the training samples, which give us a stronger baseline. To build the Feature Bank and Label Bank at the first epoch, we use the model pre-trained only on labelled training samples. And during the testing stage, we directly construct neighborhood sub-graph by retrieving neighbors from the Feature Bank built in the training stage for each test sample. 

\subsection{Ablation Studies}
\subsubsection{Density-Aware Neighboring Aggregation}
\begin{figure}[t]
	\centering
	\includegraphics[width=1.0\linewidth]{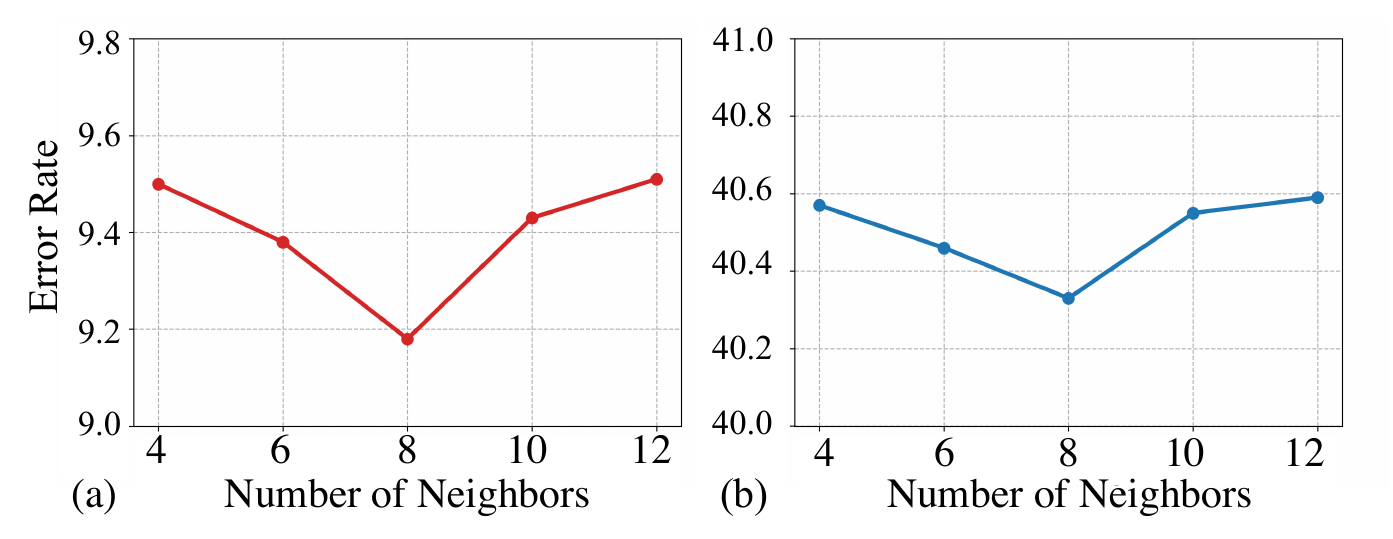}
	\caption{\small Influence of number of neighbors on CIFAR10(a) and CIFAR100(b).} \label{fig:neighbor_number_study}
\end{figure}
\begin{table}[t]
  \begin{center}
    \setlength{\tabcolsep}{3.5mm}{
    \begin{tabular}{lcccc}  
    \toprule  
   {\textbf{Method}}&  
    \multicolumn{1}{c}{CIFAR10}&\multicolumn{1}{c}{CIFAR100}\cr  
    No. of labelled images &1000&4000\cr  
    \hline \hline
    h = 0(baseline) & 10.96  & 43.34     \cr
    h = 1  & \textbf{9.18} & 40.33  \cr  
    h = 2  & 9.20 & \textbf{39.60}  \cr  
    h = 3 & 9.26  &  40.28  \cr  
    \bottomrule  
    \end{tabular} }
    \captionof{table}{\small Influence of neighbor-hops for neighborhood aggregation on CIFAR10 and CIFAR100.  } \label{tab:neighbor_hops}
     \end{center}
    \vspace{-3em}
\end{table}

\noindent  \textbf{Effectiveness of Neighborhood Aggregation.} We propose to aggregate the neighboring features to enhance the feature of the target instance, thus improving the performance of the semi-supervised image classification. To show the effectiveness of neighborhood aggregation, we conduct a baseline experiment without neighborhood aggregation and provide the comparison results in Tab.\ref{tab:Neighborhood_Aggregation}. It shows that neighborhood aggregation (``NA w/o density" and ``NA with density") can significantly improve the baseline without neighborhood aggregation (``Baseline"). To have a deeper analysis, we further visualize the learned feature embedding of neighborhood aggregation and that of the baseline on the CIFAR10 validation set in Fig.~\ref{fig:feat_visualize}, which clearly shows that incorporating neighbourhood features can help learn more discriminative feature embeddings.

\noindent  \textbf{Is density improving Neighborhood Aggregation?} When learning the aggregation weights, besides considering the feature similarity between target node and its neighbors, we believe incorporating the density information of each neighbor for aggregation weight learning is also very important based on the \textit{density-peak assumption}. To verify it, we compare the proposed density-aware neighbor aggregation (``NA with density") with the version without considering the density (``NA w/o density"). The results  in Tab.\ref{tab:Neighborhood_Aggregation} show that incorporating the density information of neighbors into the learning of aggregation weight can generally achieve superior performance. 

\noindent  \textbf{Study about Sub-Graph size.}
By experiments, we find selecting a good neighbor number $k$ and sub-graph depth $h$ (``hop") is crucial to get the best performance. First to study the influence of $k$, we only consider different number of neighbors in the first hop ($h=1$). 
The results in Fig~\ref{fig:neighbor_number_study} show that a too large or too small number of neighbors will both result in inferior results. This is because a too small number of neighbors will not get sufficient neighbouring information while a too large number of neighbors will introduce unrelated neighbours which may weaken the effectiveness of neighborhood aggregation, which is consistent with the results in~\cite{mne_iccv19}. We then study if incorporate multi-hop neighbors ($h>1$) can bring performance gain and show the results in Tab.~\ref{tab:neighbor_hops}. It can seen that incorporating two hops of neighbors can bring addition gain on CIFAR100, while yield no addition gain on CIFAR10. On the other hand, sampling the third hop of neighbors will degrade the performance on both datasets. We think it is because more unrelated samples may also be introduced as the neighbors hop increases, thus impairing the target feature. 
\vspace{-1em}
\subsubsection{Density-Ascending Path Label Propagation}
 \begin{table}[t]
  \begin{center}
    \begin{tabular}{lcccc}  
    \toprule  
   {\textbf{Method}}&  
    \multicolumn{2}{c}{CIFAR10}&\multicolumn{2}{c}{CIFAR100}\cr  
   \cmidrule(lr){2-3} \cmidrule(lr){4-5}
    No. of labelled images &1000&4000&4000&10000\cr  
    \hline \hline
    Baseline & 10.96  & 7.85 & 43.34 & 37.80     \cr
    LP-L & 9.57 & 7.37 & 40.94 & 35.94 \cr  
    LP-LU &\textbf{9.14} & \textbf{7.03} &  \textbf{40.08} & \textbf{34.72} \cr  
    \bottomrule  
    \end{tabular} 
    \captionof{table}{\small Effectiveness of Density-ascending Path-based Label Propagation on CIFAR10 and CIFAR100.  } \label{tab:Density_Ascending}
     \end{center}
    \vspace{-2em}
\end{table}

\begin{figure}[t]
	\centering
	\includegraphics[width=1.0\linewidth]{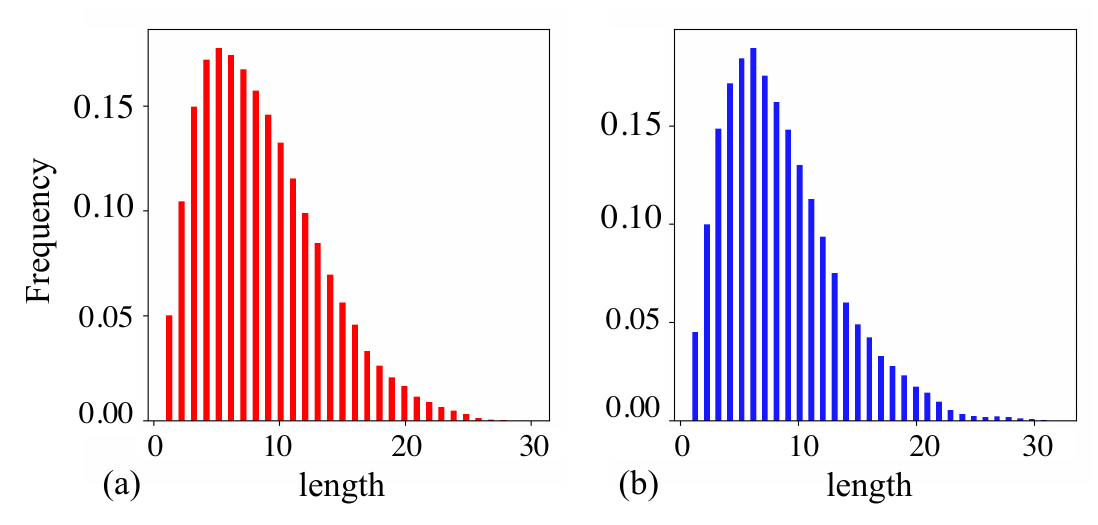}
	\caption{\small The distribution of length of density-ascending path on CIFAR10 (a) and CIFAR100 (b).} \label{fig:distribution_density_path}
\end{figure}

\begin{table}[t]
  \begin{center}
  \setlength{\tabcolsep}{3.5mm}{
    \begin{tabular}{lcccc}  
    \toprule  
   {\textbf{Method}}&  
    \multicolumn{1}{c}{CIFAR10}&\multicolumn{1}{c}{CIFAR100}\cr  
    No. of labelled images &1000&4000\cr  
    \hline \hline
    Baseline & 10.96  & 43.34      \cr
    DNA & 9.18 & 39.60 \cr  
    DPLP & 9.14 & 40.08  \cr  
    DAG(Overall framework) &\textbf{8.35}  &  \textbf{38.04} \cr  
    \bottomrule  
    \end{tabular} }
    \captionof{table}{\small  Effectiveness of overall framework on CIFAR10 and CIFAR100.} \label{tab:val_overall_framework}
     \end{center}
    \vspace{-3em}
\end{table}

\begin{table*}[ht!]
 \small
  \begin{center}
    \begin{tabular}{lcccccc}  
    \toprule  
     {\textbf{Method}}&  
    \multicolumn{4}{c}{CIFAR10}&\multicolumn{2}{c}{CIFAR100}\cr  
   \cmidrule(lr){2-5} \cmidrule(lr){6-7}
   No. of labelled images & 500 & 1000&2000&4000&4000&10000\cr  
    \hline \hline
    $\Pi$ model~\cite{laine2016temporal}& - & - & - & 12.36 $\pm~ 0.31$ & -& 39.19 $\pm ~0.36$    \cr
   TemporalEnsembling~\cite{laine2016temporal} & -& -& - & 12.16 $\pm~0.24$ & - & 38.65 $\pm~0.51$ \cr  
    MeanTeacher~\cite{tarvainen2017mean} & - & 21.55 $\pm~ 1.48$ & 15.73 $\pm~ 0.31$ & 12.31 $\pm ~0.28$& - & - \cr  
   SNTG\cite{luo2018smooth} & - & 18.41 $\pm ~0.52$& 13.64 $\pm~ 0.32$ & 9.89 $\pm ~0.34$ & - & 37.97 $\pm ~0.29$ \cr  
   SWA~\cite{athiwaratkun2018there} & - & 15.58 $\pm ~0.12$ & 11.02 $\pm ~0.23$& 9.05 $\pm~ 0.21$ & - & 33.62 $\pm ~0.54$ \cr  
   ICT \cite{ICT2019}& - & 15.48 $\pm~ 0.78$& 9.26 $\pm ~0.09$ & 7.29 $\pm ~0.02$ &- & - \cr
   DualStudent \cite{ke2019dual} & -  &14.17 $\pm ~0.38$ & 10.72 $\pm~ 0.19$& 8.89 $\pm~ 0.09$&  - & 32.77 $\pm ~0.24$ \cr  
   MixMatch\cite{berthelot2019mixmatch} &9.65 $\pm~ 0.94$ & 7.75 $\pm~ 0.32$& \textbf{7.03} $\pm~ \bf{0.15}$ & 6.24 $\pm ~0.06$ & - & - \cr  
    \hline
   TSSDL\cite{shi2018transductive} & - & 18.41 $\pm ~0.92$ & 13.54 $\pm~ 0.32$ & 9.30 $\pm~ 0.55$ &  - & - \cr  
    LP\cite{iscen2019label} & 24.02 $\pm~ 2.44$& 16.93 $\pm ~0.70$ & 13.22 $\pm~ 0.29$ & 10.61 $\pm~ 0.28$ & 43.73 $\pm ~0.20$ & 35.92 $\pm~ 0.47$ \cr  
    \hline
   DAG(Ours)  & \textbf{9.30} $\pm ~\bf{0.73}$& \textbf{7.42} $\pm ~\bf{0.41}$ & 7.16 $\pm ~0.38$& \textbf{6.13} $\pm ~\bf{0.15}$ &  \textbf{37.38} $\pm ~\bf{0.64}$ & \textbf{32.50} $\pm ~\bf{0.21}$\cr  
    \bottomrule  
    \end{tabular} 
    \captionof{table}{\small Comparison with state-of-the-art methods on CIFAR10 and CIFAR100. Average error rate and standard deviation of 5 runs with different labelled/unlabelled splits are reported. } \label{tab:sota_cifar10_cifar100}
     \end{center}
    \vspace{-2em}
\end{table*}

\begin{table}[h!]
  \small
  \begin{center}
   \setlength{\tabcolsep}{3.0mm}{
     \begin{tabular}{lll}
      \toprule 
      \textbf{Method} & 4000 & 10000 \\
      \midrule
      MeanTeacher~\cite{tarvainen2017mean} & 72.51 $\pm~0.22$ & 57.55 $\pm~1.11$ \\
       LP~\cite{iscen2019label} & 70.29 $\pm~0.81$ & 57.58 $\pm~1.47$ \\
      LP\,+\,MeanTeacher\,\cite{iscen2019label} & 72.78 $\pm~0.15$ & 57.35 $\pm~1.66$ \\
      \hline
      DAG(Ours) &\textbf{55.97} $\pm~\bf{0.62}$  & \textbf{47.28} $\pm~\bf{0.20}$  \\    
      \bottomrule
    \end{tabular}}
     \caption{\small Comparison with state-of-the-art methods on Mini-ImageNet. Average error rate and standard deviation of 3 runs with different labelled/unlabelled splits are reported.}
    \label{tab:sota_mini}
  \end{center}
  \vspace{-1cm}
\end{table}
\noindent  \textbf{Effectiveness of Density-Ascending Path.} In this part, we study the effectiveness of our proposed Density-ascending Path-based Label Propagation(DPLP) which consists of two main sequential steps: first construct a density-ascending path from each labelled sample(denoted as LP-L), then construct a density-ascending path from each remaining unlabelled sample(denoted as LP-LU). Here we study these two steps respectively in Tab.~\ref{tab:Density_Ascending}. It shows that: (i) Density-Ascending Path-based Label Propagation can significantly improve the classification accuracy; (ii) Constructing the density-ascending path only from labelled samples(LP-L) has already significantly improved the baseline; (iii) Constructing the density-ascending path from remaining unlabelled samples(LP-LU) can further bring additional gain.

\noindent  \textbf{Distribution of the length of Density-Ascending Path.} As elaborated before, our density-ascending path-based label propagation constructs the path based on the ascending density constraint and terminates according to the feature similarity constraint. Though we have already demonstrated its effectiveness, we are still curious about the distribution of the length of the density-ascending path. In Fig.~\ref{fig:distribution_density_path}, we show the distribution of the density-ascending path length on CIFAR10(1k labelled samples) and CIFAR100(4k labelled samples) respectively. We can observe that the length of density-ascending path on CIFAR10(1k labelled samples) and CIFAR100(4k labelled samples) have a similar distribution, and is mostly between in 5 and 25. Therefore, it is efficient to online perform label propagation on this density-ascending path during the training stage.
\vspace{-2em}
\subsubsection{Density Aware Graph-based SSL Framework}
In the previous subsections, we have demonstrated the effectiveness of each individual component, now we will study the effectiveness of the overall framework(DAG), $\ie$, the combination of Density-aware Neighborhood Aggregation(DNA) and Density-ascending Path-based Label Propagation(DPLP). The results on CIFAR10 and CIFAR100 are displayed in Tab.~\ref{tab:val_overall_framework}. It can be seen that DNA and DPLP are two complementary modules, and combining them can outperform the baseline by a large margin. For example, DAN and DPLP can reduce the error rate to $39.6\%$ and $40.08\%$ on CIFAR100(4k labelled samples) respectively, and their combination further reduces the error rate to $38.04\%$.

\noindent  \textbf{Discussions about computational complexity.}  The main computation of our framework comes from global graph construction which involving kNNs retrieval, yet there already exists many highly efficient nearest neighbour searching algorithms and tools. The default tool we used is Faiss\cite{JDH17}, which can perform efficient billion-scale similarity search with GPU. And for testing, we found the overhead of kNNs search is negligible compared to the feature extraction part, our test time is almost identical to the baselines.
\subsection{Comparison with state-of-the-arts}
We report the results with the state-of-the-art approaches in Tab.~\ref{tab:sota_cifar10_cifar100} and Tab.~\ref{tab:sota_mini}. To show our superiority, we consider both state-of-the-art consistency-regularization based methods~\cite{laine2016temporal, tarvainen2017mean, luo2018smooth, athiwaratkun2018there, ICT2019, ke2019dual, berthelot2019mixmatch} and pseudo-label based methods~\cite{shi2018transductive, iscen2019label} in Tab.~\ref{tab:sota_cifar10_cifar100} for CIFAR10 and CIFAR100. Among them, ICT~\cite{ICT2019} and MixMatch~\cite{berthelot2019mixmatch} both leveraged the Mixup data augmentation strategy~\cite{zhang2018mixup}, which is also used in this work. It can be seen that our method outperforms most state-of-the-art methods on CIFAR10 and CIFAR100 in terms of different numbers of labelled samples. On the more challenging Mini-ImageNet benchmark, our method achieves the best performance and records a new state-of-the-art $55.97\%$ for 4k labelled samples and $47.28\%$ for 10k labelled samples respectively, which beats latest best results \cite{iscen2019label} by 14.32\% and 10.07\%.
\section{Conclusion}
Although existing SSL methods are based on the common density-based \textit{cluster assumption} and achieve impressive results, we find three limitations exist: 1) They have not exploited density information explicitly; 2) Neighborhood information is not considered when learning the feature; 3) Existing label propagation scheme can only be done offline and difficult to be end-to-end trained; In this paper, we propose a novel and unified density-aware graph based framework for semi-supervised visual recognition. Specifically, we propose two novel density-aware modules targeting at the two key SSL components respectively, i.e., Density-aware Neighborhood Aggregation and Density-ascending Path-based Label Propagation. These two modules can be jointly trained and work in a complementary way. Experiments demonstrate our superior performance, which beats current state-of-the-art methods by a large margin.
\vspace{-1em}
\section*{Acknowledgement}
\vspace{-1em}
This work is supported by the Fundamental Research Funds for the Central Universities (WK2100330002, WK3480000005), National Key Research and Development Program of China(2018YFB0804100).

{\small
\bibliographystyle{ieee_fullname}
\bibliography{mre}
}

\end{document}